\documentclass[final]{cvpr}

\usepackage{times}
\usepackage{epsfig}
\usepackage{graphicx}
\usepackage{amsmath}
\usepackage{amssymb}
\usepackage{bbm}
\usepackage{subfigure}
\usepackage{caption}

\usepackage{times}
\usepackage{epsfig}
\usepackage{graphicx}
\usepackage{amsmath}
\usepackage{amssymb}
\usepackage{amsmath,amssymb} % define this before the line numbering.
\usepackage{color}
\usepackage{enumitem}

\usepackage{multirow}
\usepackage{makecell}

\usepackage[pagebackref=true,breaklinks=true,colorlinks,bookmarks=false]{hyperref}

\def\cvprPaperID{7608} % *** Enter the CVPR Paper ID here
\def\confYear{CVPR 2021}

\begin{document}

%%%%%%%%% TITLE
\title{Devil's in the Details: Aligning Visual Clues for Conditional Embedding in Person Re-Identification}

\author
{Fufu Yu\textsuperscript{\rm 1}, 
Xinyang Jiang\textsuperscript{\rm 1}\thanks{Fufu Yu and Xinyang Jiang contribute equally}, Yifei Gong\textsuperscript{\rm 1}, 
Shizhen Zhao\textsuperscript{\rm 3}, 
Xiaowei Guo\textsuperscript{\rm 1}, \\ 
Wei-Shi Zheng\textsuperscript{\rm 2}, 
Feng Zheng\textsuperscript{\rm 4}
Xing Sun\textsuperscript{\rm 1} \thanks{Correspondance Author: winfredsun@tencent.com} \\
\textsuperscript{\rm 1} Tencent Youtu Lab, \textsuperscript{\rm 2} Sun Yat-sen University \\
 \textsuperscript{\rm 3} Huazhong University of Science and Technology, \textsuperscript{\rm 4} Southern University of Science and Technology\\
%{\tt\small \{sevjiang, fufuyu, yifeigong, scorpioguo, garyhuang, winfredsun\}@tencent.com}\\
%{\tt\small zhaosz@hust.edu.cn}
%{\tt\small  wszheng@ieee.org@ieee.org}
% For a paper whose authors are all at the same institution,
% omit the following lines up until the closing ``}''.
% Additional authors and addresses can be added with ``\and'',
% just like the second author.
% To save space, use either the email address or home page, not both
}

\maketitle

\begin{abstract}
%In recent years, Person re-identification grows into a very important field for both academic research and industrial application. 
Although Person Re-Identification has made impressive progress, difficult cases like occlusion, change of view-point and similar clothing still bring great challenges. 
Besides overall visual features, matching and comparing detailed information is also essential for tackling these challenges. 
This paper proposes two key recognition patterns to better utilize the detail information of pedestrian images, that most of the existing methods are unable to satisfy. 
Firstly, \emph{Visual Clue Alignment} requires the model to select and align decisive regions pairs %\louis{region pairs} 
from two images for pair-wise comparison, while existing methods only align regions with predefined rules like high feature similarity or same semantic labels. 
Secondly, the \emph{Conditional Feature Embedding}  requires the overall feature of a query image to be dynamically adjusted based on the gallery image it  matches, while most of the existing methods ignore the reference images.  
%Most of the existing methods are unable to satisfy both key-point alignment and conditional feature embedding. 
By introducing novel techniques including correspondence attention module and discrepancy-based GCN, we propose an end-to-end ReID method that integrates both patterns into a unified framework, called CACE-Net ((C)lue (A)lignment and (C)onditional (E)mbedding). The experiments show that CACE-Net achieves state-of-the-art performance on three public datasets. 
%Our method first automatically selects and aligns decisive key-point pairs by a novel correspondence attention module, and then given the complex relation among key-points, it extracts conditional feature embedding with a novel discrepancy based graph convolutional network. 
%\keywords{}
\end{abstract}

\section{Introduction}
% The challenge of person reid
% Existing method 1) local feature (pcb, mgn) does not consider local matching 2) Aligned reid consider local matching but does not look at image pairs but one image
% Our method 
%weight on both image and extract pairwise feature

\begin{figure}[t]
\centering
\includegraphics[width=0.35\textwidth]{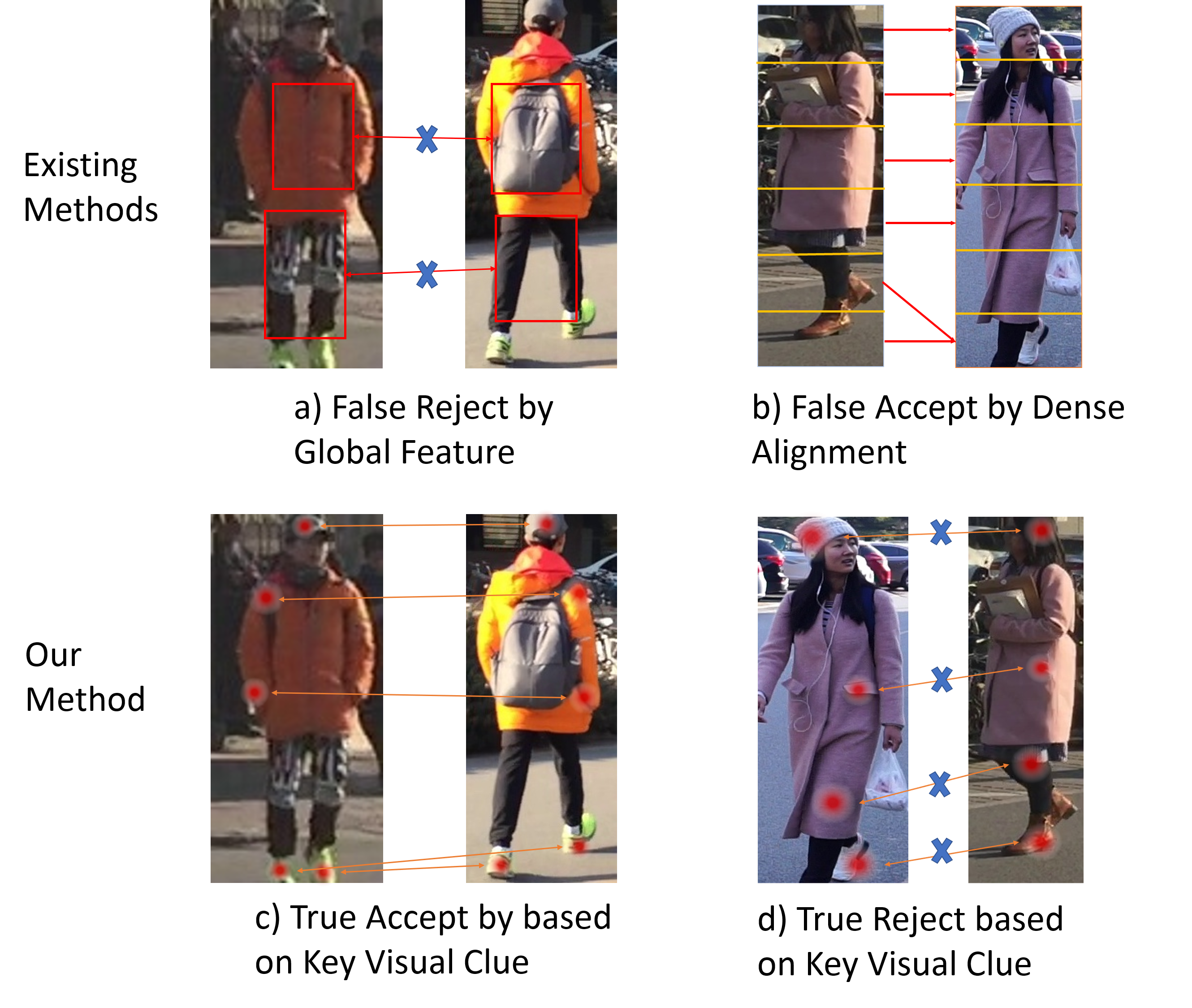} % Reduce the figure size so that it is slightly narrower than the column. Don't use precise values for figure width. This setup will avoid overfull boxes.
\caption{Visual Clue Alignment: The illustration of matching image pairs with global feature and local correspondence}
\label{fig_local_matching}
\end{figure}

Person re-identification (ReID) 
%aims at matching pedestrians across non-overlapping camera views, which 
increasingly draws attention due to its wide applications in surveillance, tracking, smart retail, \textit{etc} \cite{zheng2015scalable,Zhao_2017_ICCV,Sun_2018_ECCV}. 
Although ReID methods progress rapidly and achieve impressive performance on benchmark datasets, in practice, difficult cases like occlusion, change of view-point and similar clothing still bring great challenges. As shown in Figure \ref{fig_local_matching} a), in these cases, the overall appearance of a pedestrian may not always be reliable, and comparing the detailed information becomes essential. Thus, this paper focuses on how to effectively utilize the detailed information for matching pedestrian images. 

%This paper focuses on solving the problem of recognizing pedestrians with high view variance by conditional feature embedding. 

%In this paper, we design a novel ReID framework by analyzing human's recognition pattern on matching a pair of pedestrian images.  
Looking at how human annotators would compare the similarities between two images, we find that there are two key recognition patterns involving matching  detailed features. 
%for a human to match an image pair. 
As shown in Figure \ref{fig_local_matching} a) and b) existing methods usually compare the overall visual similarity of the entire body or densely compare the similarity of all local regions, which could be  unreliable in many hard cases. % is unreliable for matching a pair of images (e.g., ), %there is a high possibility the image pair in Figure \ref{fig_local_matching} a) will not be recognized as the same person based on overall similarity, 
On the other hand, a human annotator will select several local regions crucial for recognition, and align the selected visual clues between two images for pairwise comparison. 
For example, in Figure \ref{fig_local_matching} c), visual clue pairs including hat, shoulder, arms, and shoes are selected for comparison, and since all of these pairs have high feature similarity, one will  have high confidence to accept the images as the same person. The same logic can be applied to negative pairs, as shown in Figure \ref{fig_local_matching} c), the general appearance of the image pair is very similar, but one recognize the images pair as irrelevant by comparing visual clues including the head, legs, shoes and coat pockets. 

\begin{figure}[t]
\centering
\includegraphics[width=0.4\textwidth]{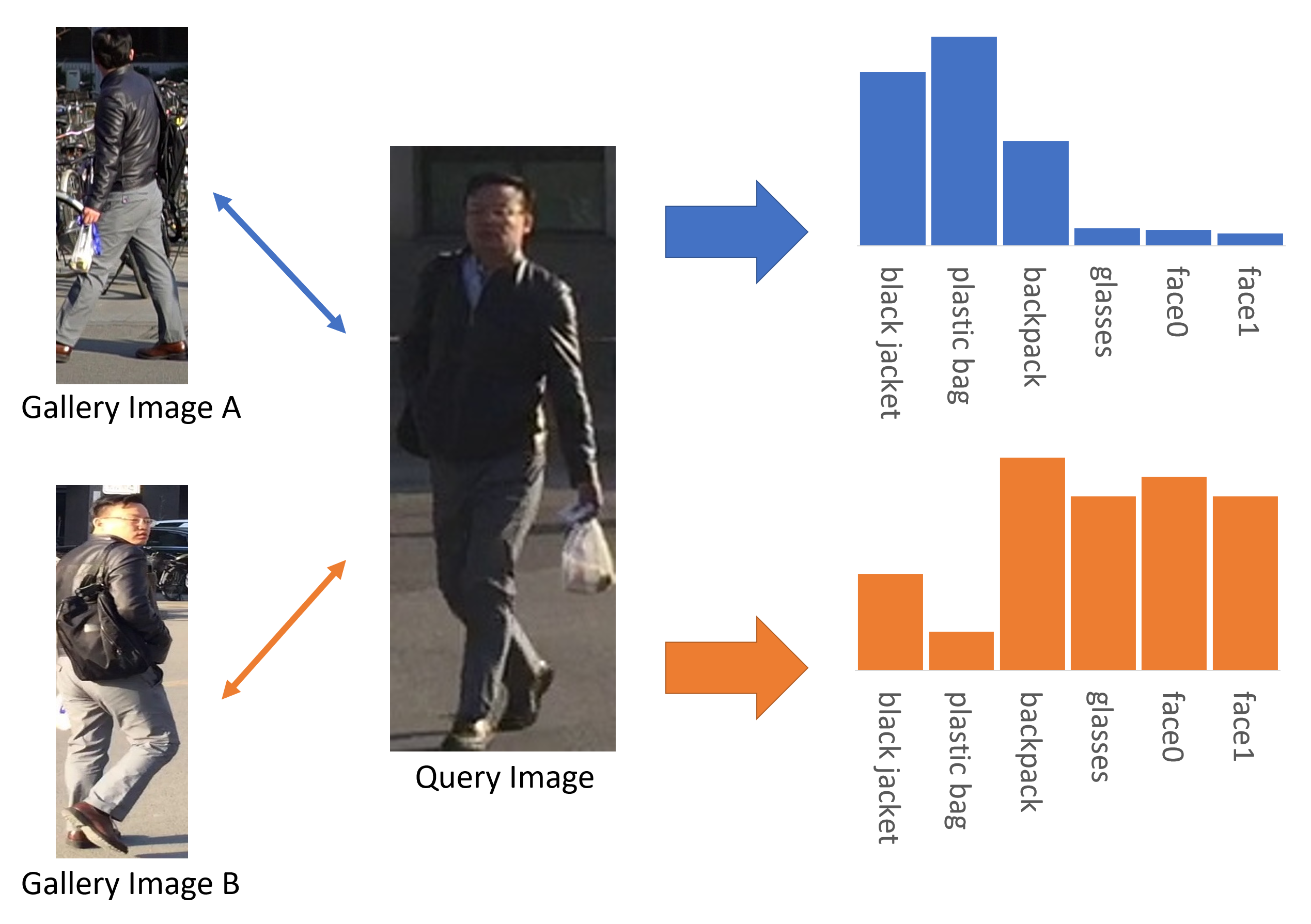} % Reduce the figure size so that it is slightly narrower than the column. Don't use precise values for figure width. This setup will avoid overfull boxes.
\caption{Conditional Feature Embedding: The illustration of changing conditional feature when comparing a image with different contextual images. }
\label{fig_conditional_feature}
\end{figure}

Secondly, for the same query image, a human annotator's attention to visual features varies drastically when matching with different gallery images. Figure \ref{fig_conditional_feature} is an intuitive example to explain this recognition pattern. For the same query image in Figure \ref{fig_conditional_feature}, the value of its feature vectors is conditioned on the gallery image it matches. In other word, different feature vectors are needed to match gallery image A and gallery image B. Since the face and glasses cannot be seen in gallery image A, the channels related to these semantics are suppressed in the corresponding feature. Similarly, when matching query image with gallery image B, channels related to the black jacket and plastic bag are suppressed. 

In conclusion, a good ReID matching model should meet two requirements: 1) Locally, decisive visual clues need to be discovered and aligned for pair-wise comparison, namely Visual Clue Alignment. 
2) Globally, the overall feature extracted from a query image should be dynamically adjusted based on the gallery image it matches, namely Conditional Feature Embedding. 

%Many existing ReID models already propose to enhance cross-view matching by adapting different local matching strategies similar to aforementioned human behaviour. Here we discuss three types of ReID model methods and exam if existing methods meets the two key requirements (i.e. conditional feature embedding and key-point alignment and selection).  

%\begin{table*}[!t]
%\centering
%\caption{The comparison of state-of-the-art ReID methods on meeting recognition patterns of key-point alignment and conditional feature embedding. }
%\label{table_requirement_compare}
%\begin{tabular}{c|c|c|c}
%\hline
%Category & Methods & Key-point Align. & Cond. Feat. Embed. \\
%\hline
%\multirow{3}*{\makecell{Local Feat. \\Learn.}}
%& Stripe:  MGN \cite{wang2018learning}, Pyramid \cite{Zheng_2019_CVPR}  & N/A & N/A \\
%& Human Parsing:  PIE \cite{zheng2019pose} & N/A & N/A \\
%& Attention: ABDNet \cite{chen2019abd} & N/A & N/A  \\
%\hline
%\multirow{2}*{\makecell{Alignment \\Based}}
%& AlignedReID \cite{Li_2018_CVPR} & Similarity-based & N/A \\
%& VPM PCB \cite{Sun_2018_ECCV},\cite{sun2019perceive} & Part Visibility & N/A \\
%\hline
%Joint Feat. Learn. & DCCs \cite{wu2018deep} & N/A & RNN-based\\
%\hline
% & Our method &  Corres. Att. & discrepancy-based GCN \\
%\hline
%\end{tabular}
%\end{table*}

%Many existing state-of-the-art methods boost the ReID performance by utilizing  detailed information. Next we examine if these methods meet above-mentioned two requirements. Table \ref{table_requirement_compare} summarizes how these methods meet the requirements. 
%As shown in Figure \ref{fig_framework}, 
Most of the existing methods do not satisfy the aforementioned two requirements. 
%Although local feature learning methods \cite{wang2018learning,li2017learning,zhao2017spindle,zheng2019pose,si2018dual} significantly boost ReID performance by learning features of local body regions, they do not conduct pairwise comparison among the regions but directly fuse all region features into a unified vector. 
Some recent alignment-based methods propose to densely align the local parts between image pair based on semantic human part labels \cite{wang2020high,sun2019perceive} or feature similarity \cite{Sun_2018_ECCV,zhang2017alignedreid,he2018deep} for pairwise comparison. 
%The overall similarity of the image pair is decided by the local similarity of the aligned part pairs. 
%The local features of the alignment-based methods are pre-extracted from the individual images and is not adjustable to different contextual images.  %and no contextual information from the the other image is considered. %Hence, alignment based methods do not achieve joint feature learning. 
However, these methods align local regions for pair-wise comparison based on pre-defined rules (e.g., regions with highest feature similarity or region with the same human part label). 
These rules in many cases are not able to discover the most decisive and discriminative visual clues. 
%.  is not flexible enough to select suitable key-point pairs for different image pairs. 
For example, as shown in Figure \ref{fig_local_matching} the importance of the aligned pairs is irrelevant to the feature similarity. In Figure \ref{fig_local_matching} b), several visual clue pairs with high visual similarity are selected, while in Figure \ref{fig_local_matching} d) region pairs with large visual difference should be selected to reject the image pair. 
Instead of predefined rules, we propose to learn a novel correspondence attention module to automatically select decisive key-point pairs based on the visual content of both images.
Furthermore, most of the existing methods extract individual features from each image and unable to learn dynamically adaptive conditional features. 
% Two methods \cite{wu2018deep,wu2018and}  learn conditional feature by RNN without clue alignment and achieve unsatisfied results. 

%Recently Wu et al. proposes a series of ReID methods to learn conditional feature embedding (i.e. DCCs \cite{wu2018deep} and Deep Spatially Multiplicative Integration \cite{wu2018and}). They learn conditional features by feeding a pair of fused global feature vectors into a RNN, but key-point alignment is not applied. Furthermore, we argue that RNN, although excels at modeling linear relation in a sequential structure, is less suitable to model the complex topological structure of local region correspondence. 

%In conclusion, to the best of our knowledge, existing methods are not able to apply both conditional feature embedding and visual clue alignment into the same model. 
As a result, we propose a novel ReID model that integrates both recognition patterns into a unified framework, called CACE-Net. 
%Our method proposes a novel graph based model to extract conditional feature embedding for image pairs based on decisive key-point pairs. 
%To match a pair of image, Siamese-GCN not only automatically selects decisive key-point pairs with a novel   but also extracts conditional local features based on contextual information from corresponding local region from another image. 
As shown in the orange box of Figure \ref{fig_pipeline}, for clue alignment, instead of pre-defined alignment rules, we propose a novel correspondence attention module to automatically select and align crucial regions between images. 
Secondly, since the region correspondence obtained by the last stage forms very complex correspondence graph, GCN is an appropriate tool to capture the the high ordered topological relation among multiple regions. As shown in grey box of Figure \ref{fig_pipeline}, our model takes the obtained region correspondence graph as inputs and extracts conditional feature embedding with a novel graph convolutional network \cite{kipf2016semi,bruna2013spectral}. 
Instead of the standard GCN that smooths the adjacent node features, a novel discrepancy-based graph convolution is proposed to obtain the feature difference between crucial regions.

%, forming an inner-image relation graph and inter-image relation graph. 

The contribution of our proposed method is listed as follows:
1) CACE-Net is able to integrate both visual clue alignment and conditional feature embedding into a unified ReID framework.  %which first automatically predicts decisive patch pairs to form a relation graph, and then extract conditional feature embedding from the graph. 
2) Instead of using a pre-defined Adjacency Matrix, our CACE-Net uses a novel correspondence attention module where the visual clues is automatically predicted and dynamically adjusted during training. 
%Our proposed Pairwise-GCN simultaneously predicts the relation graph between local patches and predicts the graph based conditional feature embedding. 
3) A novel discrepancy-based graph convolution is proposed to analyze the feature difference between adjacent graph nodes.

\section{Related Works}
\subsection{Part base methods}
Part-based models learn local features of different body parts to enhance the global ReID feature on cross-view matching. One of the most common and effective type of part-based models simply split the output feature-maps of ReID model's inter-mediate layers into several horizontal stripes and learn local features for each stripe, such as PCB \cite{Sun_2018_ECCV}, %splits feature-maps from the last layer of CNN into $6$ stripes and slightly adjusts the part partition with a Refine Part Pooling layer to solve the potential mis-alignment caused by scale change and occlusion. 
MGN \cite{wang2018learning}, Pyramid  \cite{Zheng_2019_CVPR}, RelationNet \cite{park2020relation} and VA-ReID \cite{zhu2019aware}. % proposes a pyramid structured stripe model with multiple partition branches that splits the feature-map into different numbers of stripes. 
% VA-ReID \cite{zhu2019aware} proposes a view-aware angular loss to train the part-based model, but it uses a much larger backbone network (ResNext101) compared to ours, so we do not include this method in our performance comparison.
%RelationNet \cite{park2020relation} makes each part-level feature incorporate information of other body parts to obtain
%discriminative person representations.
Another type of part-based models \cite{li2017learning, zhao2017spindle, zheng2019pose}segment human body into meaningful body parts and learn local feature for each body parts. SPReID \cite{Kalayeh_2018_CVPR} learns a human parsing branch for body part segmentation and fuses local features for different parts by weighted average pooling. DSA-ReID \cite{Zhang_2019_CVPR} proposes projects human parts into a UV space and uses this UV space branch to guide the learning of a stripe model. 

\subsection{Alignment-based methods}
On top of the local features, instead of fusing the local features directly, some methods propose to align parts from a pair of images, and match a pair of images based on the similarity of their aligned part pairs. AlignReID\cite{zhang2017alignedreid} proposes  a dynamic programming algorithm to align a stripe in the image to a stripe in another image based on their local feature similarity. DSR \cite{he2018deep} and SFR \cite{he2018recognizing} propose a sparse coding method to implicitly look for similar key-point pairs by reconstructing one image's feature map with another. VPM \cite{sun2019perceive} proposes to align the stripes from two images based on the visibility of each stripes. PGFA \cite{miao2019pose} exploits pose landmarks to align stripes. HOReID \cite{wang2020high} aligns same semantic parts for Occluded ReID  by the aid of extra key-points information. CDPM \cite{wang2019cdpm} proposes to localize and align local parts by a sliding window method. Some GAN based methods like FD-GAN \cite{ge2018fd}  proposes to align local features by directly transfer the image to the same pose and viewpoint of the target image.  

\subsection{Joint Feature Learning}
In this paper, Joint Feature Learning refers to methods that feed both images into a model simultaneously to obtain a conditional feature embedding. 
Existing methods like DCCs \cite{wu2020deep} and Deep Spatially Multiplicative Integration \cite{wu2018and} learns a RNN to iteratively encode couple features step-by-step.
% HOReID \cite{wang2020high} aligns same semantic parts for Occluded ReID  by the aid of extra key-points information that we dont't need.

\subsection{Attention-based methods}
Attention based methods are a type of State-of-the-Art ReID methods that propose to select important regions or channels of a feature-maps to form the ReID feature and discard region irrelevant to recognition such as background. Unlike our CACE-Net that selects crucial region pairs based on a pair of images, most of the existing attention based methods focus on selecting important information from individual images. Method in \cite{Zhao_2017_ICCV} proposes to predict multiple attention maps for different human parts.  HA-CNN \cite{Li_2018_CVPR} 
%and Robust ReID \cite{lawen2019attention} 
uses a Harmonious Attention module to conduct feature selection both spatially and in channel-wise for a individual image. ABDNet \cite{chen2019abd} proposes a similar spatial and channel attention with an orthogonality constraint. RGA-SC \cite{zhang2020relation} stacks the pairwise relations between single feature and all features together with the feature itself to infer the current position’s attention on spatial and channel dimensions.%DuATM \cite{si2018dual} proposes a  attention mechanism, in which both intra-sequence and inter-sequence attention strategies are used for feature refinement and feature-pair alignment, respectively.

\subsection{Graph-based methods}
Recently some methods propose to use graph-based techniques to learn more complex relationship for ReID model. However, instead of exploring the complex correlation between detailed local regions inside image pairs, most of the existing methods are based on global features. SGGNN \cite{shen2018deep} use graph to represent the relation between multiple probe images and gallery images and use a graph neural network update samples' global features with a massage passing method. Group shuffling random walk \cite{shen2018person} further extend the probe-gallery relation to gallery-gallery relationship. 

\section{Methods}
\subsection{General Framework}
Figure \ref{fig_pipeline} shows the general training workflow of CACE-Net. CACE-Net is an end-to-end learning framework containing three stages, namely individual feature embedding, visual clue alignment, and conditional feature embedding. 
At the first stage, our method extracts individual feature maps for both images. Secondly, given the feature-maps of two images, a correspondence attention module is used to select crucial region pairs both within an image and between image pair. Finally, a novel discrepancy-based GCN is used to extract conditional features from the region correspondence graph. 
The following three sub-sections elaborate on the implementation and formulation the three stages.

%The individual exam stage extract feature-map from individual images. Given the individual feature-maps of a image pair, at decisive key-point search stage a pairwise attention module is used to select decisive pixels on the feature-maps and generates a graph that builds connection between the selected pixels. At joint exam stage, a novel Siamese graph convolution network is proposed to extract conditional features for both images by considering the contextual information of the other image.  

\subsection{Individual Feature Embedding}

The individual feature embedding stage is responsible for extracting a feature-map for each individual pedestrian image. Any type of CNN backbones for person Re-identification can be applied for the individual feature extraction. 

\begin{figure*}[t]
\centering
\includegraphics[width=0.9\textwidth]{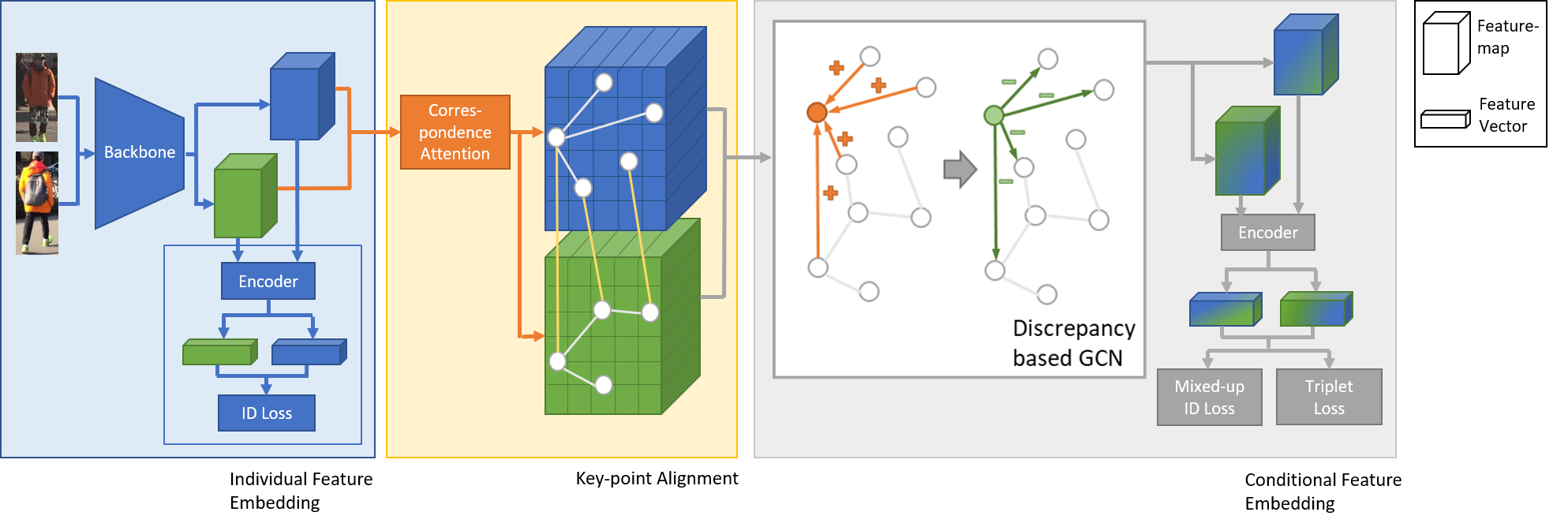} % Reduce the figure size so that it is slightly narrower than the column. Don't use precise values for figure width. This setup will avoid overfull boxes.
\caption{General Pipeline of CACE-Net. }
\label{fig_pipeline}
\end{figure*}

As shown in the blue box in Figure \ref{fig_pipeline}, to enforce the backbone network extracting good individual features, an additional training loss branch is attached to the module. 
Given a training set $\mathcal{D}=\{(I^{(u)}, y^{(u)})\}_{u=1}^N$, the input image $I^{(u)}$ is first fed into a backbone network. Then, the output feature-map is further fed into a encoder (i.e. a Global Average Pooling followed by a $1*1$ convolution layer) to obtain the individual feature vector for $I^{(u)}$, denoted as $f(I^{(u)})$. Then, %given an image with individual feature $v$ and identity $y$, 
a cross entropy based ID loss is used to train the individual feature extractor:

\begin{equation}
L_{CE}(y^{(u)}, I^{(u)}
) = \frac{1}{C}\sum_c^C \mathbf{1}(y^{(u)}=c)log(p(c|f(I^{(u)})))
\label{eq_ce}
\end{equation}
where, $p(c|f(I^{(u)})) = \frac{e^{W^T_{C}f(I^{(u)})}}{\sum_{j=1}^{K}e^{W^T_jf(I^{(u)})}}$, and $W$ is a weight matrix of a fully connect layer to classify $f(I^{(u)})$ into different identities, and $C$ is the total number of identities in the training set. 

\subsection{Visual Clue Alignment}
Visual Clue Alignment select and align crucial regions  for pairwise detailed comparison. 
We denote the feature-map extracted at individual feature embedding stage as $X \in R^{H*W*D}$, where $H$ and $W$ is the height and width of the feature-map and $D$ is the number of feature channels. $X$ is further reshaped into a two-dimensional $HW * D$ matrix denoted as $\hat{X}$. 
Given the feature-maps of two images indexed as $\hat{X}^{(u)}$ and $\hat{X}^{(v)}$, we select and align the visual clues between them by evaluating the importance of each pair of pixels, denoted as $S^{(u,v)} \in R^{HW*HW}$, where each element $S^{(u,v)}_{ij}$ in the matrix indicate the importance of an aligned pixel pair. 
Then, The selected pairs of pixels in $X$ forms a undirected graph, whose adjacent matrix $A^{(u,v)} \in R^{HW*HW}$ will be used to extract conditional features of both images with Graph Convolutional Network. 

%In a traditional GCN, the graph $A$ is usually fixed throughout the whole training process. 

%\subsubsection{Predefined Alignment Rule}、
%\subsubsection{Correspondence Attention Module}
%In CACE-Net, the visual clue alignment is conducted on the feature-maps extracted by the previous Individual Feature Embedding stage. 
One of the intuitive way to obtain the correlation between two different visual clues is to compute feature similarity. 
In the ablation study, a cosine similarity is used to evaluate the importance and correlation between a pixel pair from two feature-maps $X^{(u)}_i,X^{(v)}_j$.  
%\begin{equation}
%    S^{(u,v)}_{ij} = d(\hat{X}^{(u)}_i, \hat{X}^{(v)}_j)
%\label{eq_euclidean_attention}
%\end{equation}
Then, following a similar strategy in \cite{zhang2017alignedreid}, the crucial region pairs are selected by assigning each pixel with the pixel with  highest similarity from the other feature-map. %As a result, the adjacent matrix $A_{ij}$ between any pixels from two feature-maps are formulated as:
%\begin{equation}
%    A^{(u,v)}_{ij} = \left\{
%\begin{aligned}
%0 & & j \neq  \arg\max_k d(\hat{X}^{(u)}_i, \hat{X}^{(v)}_j)\\
%S^{(u,v)}_{ij} & & j = \arg\max_k d(\hat{X}^{(u)}_i, \hat{X}^{(v)}_j)
%\end{aligned}
%\right.  
%\end{equation}
%Noted that a continuous weight is assigned to each  instead of a binary value to indicate the connection between pixels, which gives extra information on the level of importance of each visual clue. 
%where $d$ is a predefined similarity and the euclidean distance is used in our experiments. 
Another way to build the pixel correspondence between feature-map pairs is by the human part or other semantic labels of each pixel, 
%Given a pair of pixels indexed by $i$ and $j$ with human part label $a_i$, $a_j$, the adjacent matrix is formulated as: 
%\begin{equation}
%    A^{(u)(v)}_{ij} = \left\{
%\begin{aligned}
%0 & & a_i \neq a_j  \\
%%1 & & a_i = a_j
%\end{aligned}
%\right.  
%\end{equation}
where all the corresponding human part are selected with the same importance weight. 

%\begin{figure*}[t]
%\centering
%\includegraphics[width=0.6\textwidth]{figures/relation_net.png} % Reduce the figure size so that it is slightly narrower than the column. Don't use precise values for figure width. This setup will avoid overfull boxes.
%\caption{Key-point pair search with Pairwise Attention Module. }
%\label{fig_rn}
%\end{figure*}

%\subsubsection{Correspondence Attention Module}
As discussed in Section 1, the aforementioned two types of methods for visual clue alignment and selection are based on pre-defined rules and in many cases not able to find the most decisive and discriminative region pairs. Hence, we further propose a novel correspondence attention module that automatically selects crucial visual clue pairs. 

%Given a pair of pixels $\mathbf{f}_i, \mathbf{f}_j$ in feature map, the pairwise attention model $F$ is built and learned to obtain their connection:
%\begin{equation}
%    A_{ij} = F(\mathbf{f_i}, \mathbf{f_j})
%\end{equation}

%Now we introduce our modeling of the key-point pair prediction function $F$. 
%For each pair of pixels, we propose to predict their correspondence and importance with a multi-layer perceptron. 
Our method not only focuses on finding crucial visual clues between two images, but also discovering intra-relationship between feature-map pixels within an individual image. 
The importance of pixel pairs within a feature-map  $\hat{X}^{(u)}$ is computed as follows:
%\begin{equation}
%    S_{ij} = F_{MLP}(\mathbf{f}_i\odot \mathbf{f}_j)
%\end{equation}
\begin{equation}
    S^{(u)} = \hat{X}^{(u)}W\hat{X}^{(u)\mathsf{T}}
\end{equation}
where $W$ is a diagonal parameter matrix that assigns a learn-able weight for each feature-map channel. 

Similarly, given two images whose feature-map is denoted as $X^{(u)}$ and $X^{(v)}$, the inter-image importance is computed as follows:
\begin{equation}
    S'^{(u,v)} = X^{(u)}W'X^{(v)\mathsf{T}}
\end{equation}
%\begin{equation}
%    S'_{st} = F_{MLP}'([\mathbf{f}_s, \mathbf{f}_t]) + F_{MLP}'([\mathbf{f}_s, \mathbf{f}_t]). 
%\end{equation}

We combine the importance weight of both intra-image and inter-image visual clues, and select the crucial pairs based on the importance matrix as follow: 
\begin{equation}
\begin{aligned}
A^{(u,v)}=ReLU\left(\begin{array}{cc}S^{(u)} &S'^{(u,v)}\\S'^{(v,u)}&S^{(v)}\end{array}\right)
\end{aligned}
\end{equation}
where a ReLU activation is used to select the regions pairs with positive importance weights. 

\subsection{Conditional Feature Embedding}

Given a pair of images $(I^{(u)}, I^{(v)})$, the conditional feature embedding of $I^{(u)}$ should be dynamically adjusted based on $I^{(v)}$, denoted as $f_{cond}(I_i| I_j)$. 

In order to fully exploit the detailed information in both images, we propose to extract the conditional feature embedding based on the crucial visual clue pairs between $I^{(u)}$ and $I^{(v)}$.  
Given the alignment graph $A^{(u,v)}$ of visual clues predicted by the correspondence attention module, 
%denoted with a adjacent matrix $A$, 
we propose a novel discrepancy-based GCN to encode the complex graph structured contextual information into conditional feature vectors. 

\subsubsection{Discrepancy-based GCN}
In this paper, we choose a spectral based GCN \cite{kipf2016semi} to extract conditional features. 
Given the $i$-th pixel of an image's feature-map, denoted as $\hat{X}_i^{(u)}$ (i.e. the node feature), the graph convolution operation is formulated as follows: 
%\begin{equation}
%    g_{\theta} * x = Ug_{\theta}U^{T}\hat{X}^{(u)}_i
%\end{equation}
%where $g_{\theta}$ is a function of the eigen-value of the normalized graph Laplacian $L = L - D^{-\frac{1}{2}}AD^{\frac{1}{2}}$, $U$ is the matrix of the eigen-vectors of $L$. This equation is computationally expensive due to the need of eigen decomposition of $L$. As a result, a K-order Chebyshev polynomials is used to approximate the convolution operation:
%\begin{equation}
%    g_{\theta'} * X_i^{(u)} = \sum_{k=0}^K\theta_k'T_k(\tilde{L})X_i^{(u)} 
%\label{eq_chev_k_gcn}
%\end{equation}
%where $\tilde{L}=\frac{2}{\lambda_{max}}L-I$ with $L = L - D^{-\frac{1}{2}}AD^{\frac{1}{2}}$ and $\lambda_{max}$ the maximum of the eigen value of $L$;  $T_k(\tilde{L})=2\tilde{L}T_{k-1}(\tilde{L})-T_{k-2}(\tilde{L})$, with $T_0(\tilde{L})=1$ and $T_1(\tilde{L})=\tilde{L}$.
%For further simplification,  GCN sets $K$ to $1$ and $\lambda_{max}$ is set to a fixed value $2$. Thus equation \ref{eq_chev_k_gcn} is simplified to:
\begin{equation}
\begin{aligned}
    g_{\theta'} * X_i^{(u)} &= \theta_0'X_i^{(u)} + \theta_1'(L-I)X_i^{(u)} \\
    &= \theta_0'X_i^{(u)} - \theta_1'D^{-\frac{1}{2}}A^{(u,v)}D^{\frac{1}{2}}X_i^{(u)}
\end{aligned}
\end{equation}
where $L$ is the Laplacian matrix of the adjacent matrix $A^{(u,v)}$

To further decrease the number of learn-able parameter, common GCN sets $\theta = \theta_0' = -\theta_1$, which leads to following expression:
\begin{equation}
    g_{\theta} * X_i^{(u)} = \theta (I +  D^{-\frac{1}{2}}A^{(u,v)}D^{\frac{1}{2}})X_i^{(u)}
\end{equation}
%As analyzed in \cite{li2018deeper}, 
By taking a closer look at this equation, we will find that this operation is essentially a weighted average or smooth operation over current node feature itself and its connected neighbours. However, this is not what we want for our graph convolution. Under the setting of the Re-identification, instead of smoothing the value between connected nodes, we require the model to obtain the features difference between the aligned crucial regions. Thus, We propose a novel graph convolution operation, that instead of letting $\theta_0' = -\theta_1$, sets $\theta = \theta_0' = \theta_1'$, which leads to following graph convolution operation:
\begin{equation}
\begin{aligned}
    g_{\theta} * X_i^{(u)} &= \theta (I -  D^{-\frac{1}{2}}A^{(u,v)}D^{\frac{1}{2}})X_i^{(u)} \\
    &= \theta D^{-\frac{1}{2}}L D^{\frac{1}{2}}X_i^{(u)}
\label{eq_gcn_laplacian}
\end{aligned}
\end{equation}
From Eq. \ref{eq_gcn_laplacian},  we can see that for our new graph convolution, the coefficient of $g_{\theta}$ becomes the normalized graph Laplacian matrix, which is equivalent to computing a secondary gradient of the node feature. Hence this convolution is able to obtain the level of feature change between adjacent nodes. 

\subsubsection{Mixed up ID Loss}
As shown in Figure \ref{fig_pipeline}, given the feature-maps of image pair $(I^{(u)}, I^{(v)})$ and the adjacent matrix $A^{(u,v)}$ generated from the feature-map pair, we obtain the conditional feature-map by applying the graph convolution described in Eq. \ref{eq_gcn_laplacian}. The outputs are a pair of conditional feature-map with the same size of the input feature-map. Similar to the individual exam stage, the conditional feature-maps are then fed into a feature encoder consisting of a Global Average Pooling layer and a $1*1$ convolution layer for dimension reduction to obtain the encoded conditional feature vectors . 

In the training process, we use both triplet loss and cross entropy loss as the supervised signals. In every training iteration, we sample $P$ identities from the training set and for each identity in the training set, we sample $M$ samples. %We denote the conditional feature vector extracted 
For triplet loss, we use % triplet loss that is almost the same as 
a common hard triplet loss for person ReID \cite{hermans2017defense}.
%\begin{equation}
%\begin{align}
%    L_{triplet} &= \sum_{i=1}^P\sum_{a=1}^M\Big(m + \max_{i=1}^{M-1}d\big(f_{coup}(I_a|I_i\big), f_{coup}(I_i|I_a)) \\
%    & - \min_{j\not=i}d\big(f_{coup}(I_a|I_j), f_{coup}(I_j|I_a)\big)\Big)
%\end{align}
%\end{equation}
%Training ReID model with merely triplet loss usually causes over-fitting on a very small sub-set of hard samples. Hence an extra cross-entropy loss is required. 
As for the cross-entropy loss, since the conditional feature $f_{cond}(I^{(u)}|I^{(v)})$ is extracted based on the information from both $I^{u}$ and $I^{(v)}$, the identity labels from both image should be used to supervised the feature extraction. Instead of the common cross entropy loss, we propose a mix-up cross entropy loss specifically for training conditional feature vector. Given a mini-batch containing $PM$ images, %image $I_i$ and its conditional feature conditioned on $I_j$ $f_{cond}(I_i|I_j)$, 
the mix-up cross-entropy loss is formulated as:
\begin{equation}
\begin{aligned}
L_{mix-up} &= \sum_{u = 1}^{PM}\sum_{v=1}^{PM} \alpha L_{CE}(y^{(u)}, f_{cond}(I^{(u)}|I^{(v)}))\\
&+ (1-\alpha) L_{CE}(y^{(v)}, f_{cond}(I^{(u)}|I^{(v)}))
\end{aligned}
\end{equation}
where $L_{CE}$ is the softmax and cross entropy loss shown in Eq.\ref{eq_ce} . 

\subsection{Model Inference}
Like most of the ReID method, the model is evaluated as an image retrieval task. Given a query image set containing $N_q$ images %$\{I_{q0}, I_{q1}, ... I_{qN_q}\}$ 
and a gallery set containing $N_g$ images,  
%$\{I_{g0}, I_{g1}, ..., I_{gN_g}\}$
we need to retrieve the images with the same identity of $I_q$. Our method obtains the similarity between two images with both individual features and conditional features.  %based on the feature similarity between the query image and gallery images.

%The proposed Siamese-GCN %extracts both individual features for single images and conditional features for image pairs,and 
%The computational complexity of the conditional feature extraction of  Siamese-GCN is $O(N_qN_g)$, which is very expensive for a large gallery. 
We first extract the individual features
%-maps with the backbone and feature vector with the feature encoder in individual exam stage 
for all images in query and gallery set, with computational complexity of $O(N_q + N_g)$. Then, for each query, we first sort the gallery images based on the similarity of the individual features. After that, the feature-maps of query image and the top-K images in sorted gallery forms $K$ feature-map pairs, which are fed into the key-point alignment stage and conditional feature embedding stage to obtain conditional features, with computational complexity of $O(N_qK)$. Finally, the top-K gallery images are sorted once more by the similarity of coupled features, forming the final ranking result. Compared to the individual feature extraction,  much fewer computing operations are needed to obtain conditional feature embedding, so the entire computation cost of CACE-Net is very close to normal ReID method. 

\section{Experiments}

In this section we propose the performance comparison of CACE-Net with the state-of-the-art methods and ablation study of different components in CACE-Net. 

\subsection{Datasets }
Our experiments are conducted on three widely used ReID benchmark datasets. 

\noindent \textbf{Market-1501} \cite{zheng2015scalable}
dataset contains 32,668 person images of 1,501 identities captured by six cameras. Training set is composed of 12,936 images of 751 identities while testing data is composed of the other images of 750 identities. %In addition, 2,793 distractors also exist in testing data. 

\noindent \textbf{MSMT-17} \cite{wei2018person}
dataset contains 124,068 person images of 4,101 identities captured by 15 cameras (12 outdoor, 3 indoor). Training set is composed of 30,248 images of 1,041 identities while testing data is composed of the other images of 3060 identities.  

%\vspace{-0.2cm}
\noindent \textbf{DukeMTMC-reID} \cite{ristani2016performance}
dataset contains 36,411 person images of 
1,404 identities captured by eight cameras. They are randomly divided, with 702 identities as the training set and the remaining 702 identities as the testing set. In the testing set, for each ID in each camera, one image is picked for the query set while the rest remain for the gallery set.  

\noindent \textbf{Occluded-DukeMTMC} re-splits the DukeMTMC-reID dataset to generate the new Occluded-DukeMTMC dataset. All query images and $10\%$ gallery images in the new dataset are occluded person images.

\subsection{Implementation Detail}
The input images are resized into $384 \times 128$. In training stage, we set batch size to be 16 by sampling 4 identities and 4 images per identity. The ResNet-50 \cite{he2016deep} model pretrained on ImageNet is used as the backbone network. Some common data augmentation strategies including horizontal flipping, random cropping,  random erasing \cite{zhong2017random} (with a probability of 0.5) are used. We adopt Gradient Descent optimizer to train our model and set weight decay $5 \times 10^{-4}$. The total number of epoch is 80. The learning rate is initialized to $6.25 \times 10^{-3}$ and is decayed by cosine method until it equals to 0. At the beginning, we warm up the models for 5 epochs and the learning rate grows linearly from 0 to $6.25 \times 10^{-3}$.

\subsection{Ablation Study}

In this sub-section, we report the evaluation results of the influence of different components and hyper-parameters of our method. 

\subsubsection{Influence of Model Components}
Table \ref{table_stage} shows the influence of the three stages of CACE-Net  (i.e. individual feature embedding, visual clue alignment and conditional feature embedding). Following methods are compared:
\begin{itemize}[leftmargin=*]
\setlength\itemsep{-5pt}
    \item \textbf{Individual Feature Embedding}.  This is a baseline method using only the individual feature vector extracted by the encoder in the individual exam stage. We observe that this method achieves lowest performance. 
    \item \textbf{Visual Clue Alignment}. The feature-maps extracted by the individual stage are further fed into the Visual Clue Alignment stage to get correspondent pixel pairs between two images. Instead of applying GCN, we directly compute the average feature similarity of the cross-image key-point pairs as the overall similarity of the two images. Table \ref{table_stage} shows an extra key-point alignment improves the baseline model by more than $1$ percentage point in terms of MAP. %, which verify the effectiveness of selecting and aligning the image pair's local key-points. 
    \item \textbf{Individual-GCN}. %The adjacent matrix and individual feature-maps are fed into Siamese-GCN to obtain conditional features. 
    All three stages are performed, but all cross image connections in the adjacent matrix are discarded and only intra-frame relations are considered. As show in Table \ref{table_stage}, intra-frame relation based GCN gives around $1$ percent improvement in terms of MAP, which indicates that extracting features based on local information and correlation inside individual image can boost ReID performance.  
    \item \textbf{Pairwise-GCN (Normal)}. Pairwise-GCN considers both intra-frame relation and inter-frame relation in GCN. Here the common GCN that uses a smooth operation on adjacent node is applied and we observe that normal GCN does not achieve performance improvement, which indicates that normal GCN is not suitable for extracting conditional features for ReID. 
    \item \textbf{Pairwise-GCN (Discrepancy-based)}. Our novel discrepancy-based GCN is used and achieves obvious improvement compared to normal GCN. Furthermore, adding inter-frame relation between two images outperforms individual-GCN by a clear margin, showing the effectiveness of extracting conditional features based on local correlation between image pairs. 
\end{itemize}

\begin{figure}[t]
\centering
\includegraphics[width=0.3\textwidth]{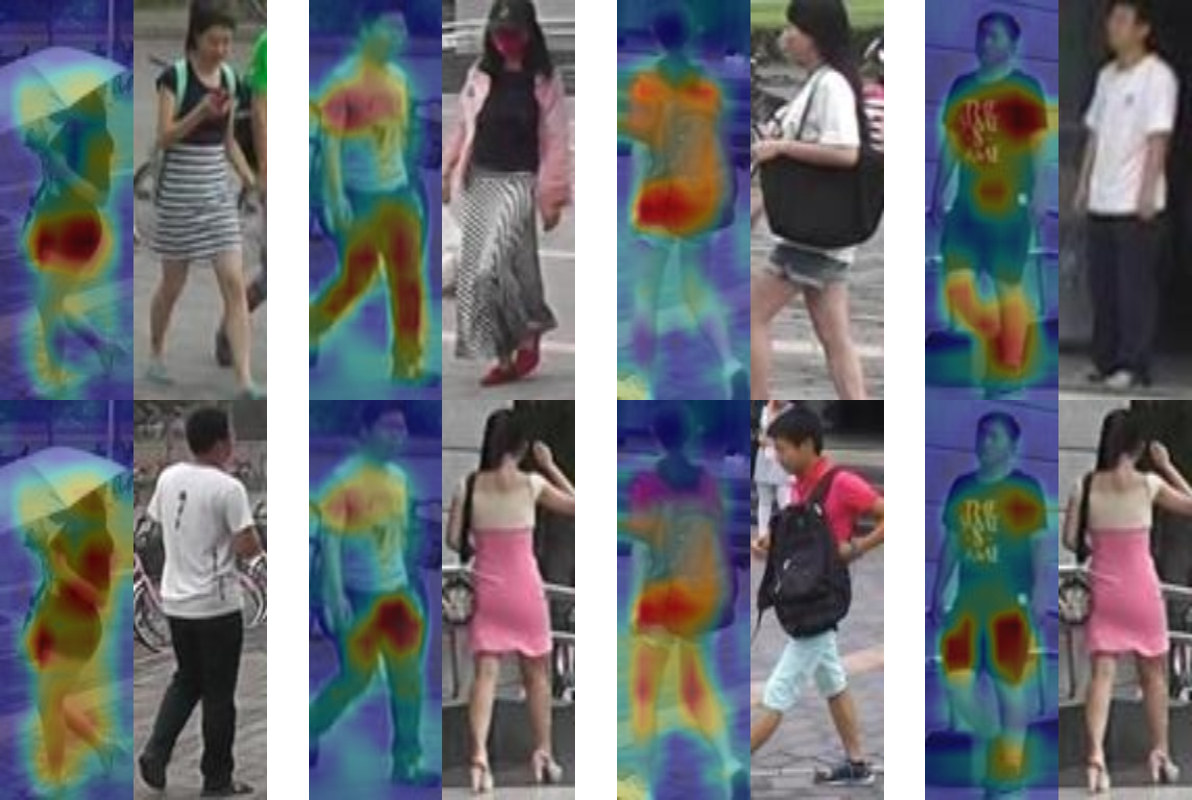} % Reduce the figure size so that it is slightly narrower than the column. Don't use precise values for figure width. This setup will avoid overfull boxes.
\caption{The visualization of Conditional Feature Embedding. The feature-map difference of the same query when matching with different images is shown. For each column, the images on the left are different feature-maps of the same query image, and the images on the right are the different gallery images the query matches. The query's feature-map is dynamically adjusted based on the gallery image.  % where similarity based alignment focuses on high similarity key-point pairs and correspondence attention selects decisive key-pairs of both high similarity and high difference. 
}
\label{fig_vis_featuremap}
\end{figure}

We visualize the conditional feature-map obtained by CACE-NET in Figure \ref{fig_vis_featuremap}. The feature-map is visualized by computing the  L2 norm of each feature in the feature-map. As shown in Figure \ref{fig_vis_featuremap}, the feature-map of the query image (the left image in each column) changes drastically based on the gallery image it matches (the right image in each column). For example, the feature-map will have higher activation on the lower body on column 1 row 2, column 2 row 1 and column 3 row 1, because the lower body has more distinctive features to tell the query image and the corresponding gallery image apart. This visualization results prove that CACE-NET is able to dynamically adjust the feature embedding based on the image it matches, i.e. the conditional feature embedding. 

\begin{table}[!t]
% p{3cm}<{\centering}
\centering
\footnotesize
\caption{Performance (\%) comparisons of three stages (Individual Feature Embedding, Visual Clue Alignment and Conditional Embedding) on DukeMTMC.} 
\label{table_stage}
\begin{tabular}{lll}
\hline
Method & mAP & Rank-1 \\
\hline
Individual Feature Embedding & 77.03 & 87.84\\
 + Visual Clue Alignment & 78.29 & 89.68  \ \\
 + Individual-GCN & 79.19 & 89.77  \\
 + Pairwise-GCN (Normal) & 78.58  & 89.77 \\
 + Pairwise-GCN (Discrepancy-based) & \textbf{81.29} & \textbf{90.89} \\
\hline
\end{tabular}
\end{table}

\begin{table}[t]
% p{3cm}<{\centering}
\centering
\footnotesize
\caption{Performance (\%) comparisons of different alignment strategy for graph generation in CACE-Net on DukeMTMC.} 
\label{table_alignment}
\begin{tabular}{llll}
\hline
Method  &mAP&Rank-1\\ %&& Rank-5\\
\hline
No Alignment (Baseline) & 77.03 & 87.84 \\
Part Alignment & 78.39 & 89.45 \\
Top-K Similarity & 79.30 & 89.72\\
Fully Connect & 80.19 & 89.90 \\
Semantic Alignment & 80.64 & 90.08\\
%Pairwise Attention (intra-image) & 88.80 & 95.67& 98.16  \\
Correspondence Attention  &\textbf{81.29} & \textbf{90.89} \\
\hline
\end{tabular}
\end{table}

\begin{table}[t]
\footnotesize
% p{3cm}<{\centering}
\centering
\caption{Performance (\%) comparisons of different $\alpha$ in  mix-up id loss on Market1501.} 
\label{table_mixup}
\begin{tabular}{lll}
\hline
Method &mAP&Rank-1\\
\hline
Cross Entropy (Baseline) & 88.57 & 94.74 \\ %& 98.13 \\
Mix-up $\alpha=0.95$  & 89.44 & 95.75 \\ %& 98.31\\
Mix-up $\alpha=0.9$  & \textbf{90.30} & \textbf{95.96} \\ %&& \textbf{98.28}\\
Mix-up $\alpha=0.8$  & 89.56 & 95.84 \\ %& 98.22 \\
Mix-up $\alpha=0.7$  & 88.72 & 95.43 \\ %& 98.16 \\
Mix-up $\alpha=0.6$  & 85.67 & 94.83 \\ %& 97.65 \\
\hline
\end{tabular}
\end{table}

\begin{table*}[t]
% p{3cm}<{\centering}
\centering
\footnotesize
\caption{Performance (\%) comparisons to the state-of-the-art results on Market-1501, DukeMTMC-reID and MSMT-17. Our proposed CACE-Net outperforms the state-of-the-art methods. } 
\label{table_sota}
\begin{tabular}{cllcccccccc}
\hline
\multirow{2}*{Category} & \multirow{2}*{Method} & \multicolumn{2}{c}{Market-1501} & \multicolumn{2}{c}{DukeMTMC-reID}& \multicolumn{2}{c}{MSMT-17} \\
\cline{3-8}
& & {mAP}&{Rank-1}&{mAP}&{Rank-1}&{mAP}&{Rank-1}\\
\hline

\multirow{4}*{Part-based}& PCB (ECCV 2018) \cite{Sun_2018_ECCV} &  77.4 & 92.3  & 66.1 & 81.7 & - & - \\
& MGN (ACM MM 2018) \cite{wang2018learning}  & 86.9 & 95.7 & 78.4 & 88.7 & - & - \\ 
& Pyramid (CVPR 2019) \cite{Zheng_2019_CVPR}  & 88.2 & 95.7 & \underline{79.0} & 89.0 & - & -\\
& RelationNet (AAAI 2020) \cite{park2020relation}  & \underline{88.9} & 95.2 & 78.6 & 89.7 & - & -\\
%& OSNet \cite{zhou2019omni}  & 84.9 & 94.8 & 73.5 & 88.6 & 52.9 & 78.7\\
%& SPReID \cite{Kalayeh_2018_CVPR} & 83.36 & 93.68 & 73.34 &  85.95 & - & - \\
%& DSA-reID (CVPR 2019) \cite{Zhang_2019_CVPR} & 87.6 & 95.7 & 74.3 & 86.2 & - & - \\
\hline

\multirow{4}*{Alignment}
%& PCB+RPP \cite{Sun_2018_ECCV} & 81.6 & 93.8  & 69.2 & 83.3 & - & - \\
& AlignedReID (ECCV 2018)  \cite{Sun_2018_ECCV} &  79.3 & 91.8 & - & - & - & -\\
& FD-GAN (NIPS 2018) \cite{ge2018fd} &  77.7 & 90.5 & 64.5 & 80.0 & - & -\\
& VPM (CVPR 2019) \cite{sun2019perceive} & 80.8 & 93.0 & - & - & - & - \\
% & SAN (AAAI 2020) \cite{jin2020semantics} & 88.0& \textbf{96.1} & 75.5 & 87.9 & 55.7 & 79.2 \\
&  HOReID (CVPR 2020) \cite{wang2020high}& 84.9  & 94.2 & 75.6 & 86.9 & - & -  \\
\hline

\multirow{3}*{Attention} 
%& HA-CNN (CVPR 2018) \cite{Li_2018_CVPR} & 75.7 & 91.2 & 63.8 & 80.5 & - & - \\
& ABD-Net (ICCV 2019) \cite{chen2019abd} & 88.28 & 95.6  & 78.59 & 89.0 & \underline{60.8} & 82.3  \\
& RGA-SC (CVPR 2020) \cite{zhang2020relation} & 88.4 & \textbf{96.1} & - & - & 57.5 & 80.3 \\
& SCSN (CVPR 2020) \cite{chen2020salience} & 88.5 & 95.7 & \underline{79.0} & \textbf{91.0} & 58.5 & \textbf{83.8} \\
%& Robust ReID \cite{lawen2019attention} & 89.7 & 95.6 & 80.3 & 89.8 & - & -  \\
\hline

\multirow{2}*{Joint Learning}
&  SMI (PR 2018) \cite{wu2018and} & 65.25  & 86.15 & - & - & - & - \\
& DCCs (TNNLS 2020) \cite{wu2020deep} & 71.1  & 88.4 & 59.2 & 80.3 & - & - \\
%& ABD-Net \cite{chen2019abd} & 88.28 & 95.60  & 78.59 & 89.00 & 60.80 & 82.30  \\

%& Deep-CRF \cite{Chen_2018_CVPR} & 81.60 & 93.50 & 97.7 & 69.5 & 84.9 &  92.3 \\
\hline
\multirow{2}*{Graph-based}
&  Group-shuffling (CVPR 2018) \cite{shen2018deep} & 82.5  & 92.7 & 66.4 & 80.7 & - & - \\
&  SGGNN (ECCV 2018) \cite{shen2018person} & 82.8  & 92.3 & 68.2 & 81.1 & - & - \\

\hline
%& RTRB \cite{ro2019rtrb} & 79.9 &  92.5 & - & 70.2 & 85.2 & - \\
%& LITM+GHIS \cite{zhang2019learning} & 83.9 & 93.9 & - & 74.5 & 85.9 & - \\
%--- & & & & & & & & & \\

\multirow{1}*{\textbf{CACE-Net}} &  \textbf{CACE-Net} & \textbf{90.3} & \underline{95.96} & \textbf{81.29} & \underline{90.89} & \textbf{62.0} & \underline{83.54}\\
 %& \textbf{Ours+reranking} & \textbf{95.43} & \textbf{96.79} & \textbf{98.31} & \textbf{91.82} & \textbf{93.85} & \textbf{96.50} \\
%\multirow{5}*{This work} & VOAL(prediction)  & 89.84 & 95.58 & 98.40 & 81.05 & 90.75 & 95.65 \\ 
% & VOAL  & 89.97 & 95.87 & 98.57 & 81.48 & 91.11 & 95.38 \\ 
% & VOAL+RR  & 95.09 & 96.32 & 98.19 & 90.66 & 92.46 & 96.23 \\
% & VOAL+local  & 91.70 & 96.23 & 98.69 & 84.51 & 91.61 & 96.23 \\ 
% & VOAL+local+RR  & 95.43 & 96.79 & 98.31 & 91.82 & 93.85 & 96.50 \\ 
%Ours  & This work & 9.86 & 52.64 & 89.64 & 95.64 & 98.28 & 80.19 & 89.63 & 95.33 \\ 
%Ours+RR  & This work & - & - & 94.45 & 96.05 & 97.98 & 91.24 & 92.64 & 96.05 \\
\hline
\end{tabular}
\end{table*}

\begin{figure}[t]
\centering
\includegraphics[width=0.34\textwidth]{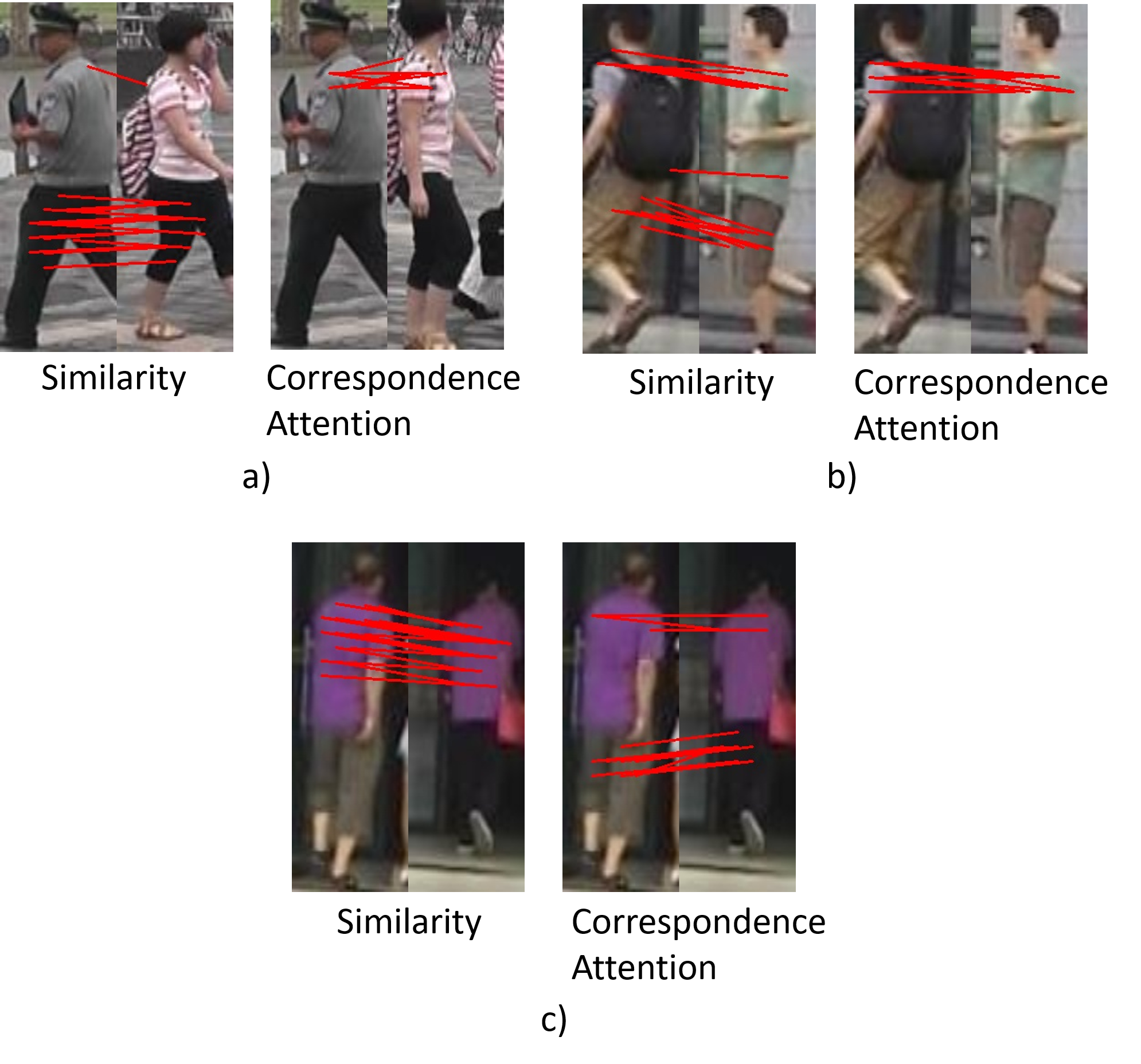} % Reduce the figure size so that it is slightly narrower than the column. Don't use precise values for figure width. This setup will avoid overfull boxes.
\caption{The cross-image alignment result obtained by similarity-based alignment and correspondence attention. The red lines denote a visual clue correspondence between two images. % where similarity based alignment focuses on high similarity key-point pairs and correspondence attention selects decisive key-pairs of both high similarity and high difference. 
}
\label{fig_vis_align}
\end{figure}

\subsubsection{Influence of Alignment Strategy}

In order to prove the advantage of using automatically learned neural network to predict the correspondence between pixel in the feature-map pairs, Table \ref{table_alignment} compares the influence of different alignment strategies to our CACE-Net method. Following alignment strategy is evaluated:
\begin{itemize}[leftmargin=*]
    \setlength\itemsep{-5pt}
    %\vspace{-3pt}
    \item \textbf{No Alignment}: The baseline method uses only feature vector from individual exam stage where no connection exists between any pixels. 
    %\vspace{-3pt}
    \item \textbf{Part-based Alignment}: Similar to the part-based model like PCB \cite{Sun_2018_ECCV}, two pixels at the same location of the images are aligned. As shown in Table \ref{table_alignment}, part alignment strategy outperforms the baseline method but achieve lower performance than correspondence attention, because it is not robust to scale changes and unable to rule out non-decisive pairs. 
    %\vspace{-3pt}
    \item \textbf{Fully Connect}: A fully connected adjacent matrix is applied where all pixels are correspondent. In this way, the contextual information for all other pixels are explored when extracting a conditional feature. This method outperforms the baseline but does not achieve the best performance,  because too much redundant contextual information is involved in a fully connected graph. 
    %\vspace{-3pt}
    \item \textbf{Similarity-based Alignment}: Similar to AlignedReID \cite{zhang2017alignedreid}, this method selects the most similar pixel as each pixel's correspondent neighbour. It does not perform as well as our method because as discussed in section 1, the predefined similarity-based alignment strategy is able to find the most decisive and discriminative region pairs. %For example, key-point pairs with low similarity to distinguish different identities may be needed. 
    %\vspace{-3pt}
    \item \textbf{Semantic Alignment}: Pixels lie in the same semantic body part are connected as neighbours whose distances are 1, unlike soft distances in our correspondence attention. Despite the help of extra human parsing network, this method is inferior to our correspondence  attention.
    %\vspace{-3pt}
    \item \textbf{Correspondence Attention Module}: With a correspondence attention module,  CACE-Net is able to build a more crucial correspondence compared to pre-defined rules and achieves the best performance. %Furthermore, we observe that introducing both intra-image and inter-image correspondence effectively boosts the performance. 
\end{itemize}

In Figure \ref{fig_vis_align} we compare the visualization results of similarity-based alignment and correspondence attention, where crucial region pairs with highest scores are visualized. We observe that similarity-based alignment that  only aligns region pairs with high visual similarity, causing mis-matching images with similar local parts (e.g.,  the lower body in Figure \ref{fig_vis_align} a,b and upper-body in Figure \ref{fig_vis_align} c). On the other hand, our correspondence attention disregards the visual similarity and is able to focus on correct decisive clues to reject image pairs (e.g. the shoulder area in Figure \ref{fig_vis_align}). 

\begin{figure}[t]
\centering
\includegraphics[width=0.4\textwidth]{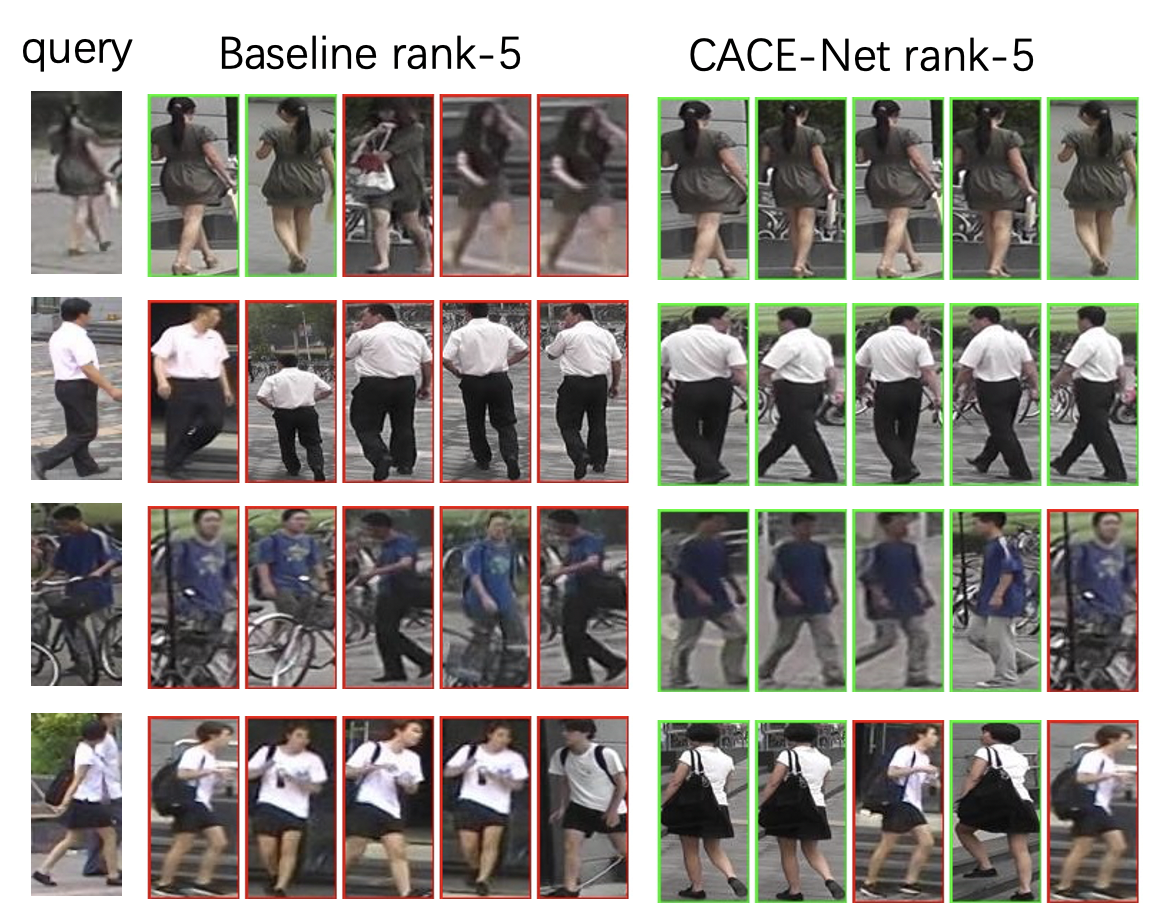} % Reduce the figure size so that it is slightly narrower than the column. Don't use precise values for figure width. This setup will avoid overfull boxes.
\caption{The comparison of retrieval results of baseline method and CACE-Net. % where similarity based alignment focuses on high similarity key-point pairs and correspondence attention selects decisive key-pairs of both high similarity and high difference. 
}
\label{fig_vis_ranklist}
\end{figure}

Figure \ref{fig_vis_ranklist} compares the retrieval results of baseline method and CACE-Net. We observe that in hard cases like similar clothing or view-point variance show in Figure \ref{fig_vis_ranklist}, CACE-Net is able to achieve much better results. It proves that even for images with extremely similar general visual appearance,  CACE-Net is able to discover and compare crucial visual clues, such as the hair style in Figure\ref{fig_vis_ranklist} row 1 and row 2, pattern on the t-shirt in row 3 and shoes in row 4. 

\subsubsection{Influence of Mix-up ID Loss}
Table \ref{table_mixup} shows the influence of our customized Mix-up Loss for conditional feature embedding. As shown in Table \ref{table_mixup}, with the right hyper-parameter $\alpha$, our proposed Mix-up Loss significantly outperforms common cross entropy loss. We also observe the influence of the hyper-parameter $\alpha$ to the model performance, where CACE-Net achieves the best performance when $\alpha$ is set to $0.9$, which shows that  conditional feature contains only small amount of information from contextual image compared to the target image.

\subsection{Comparison with the State-of-the-Art}

\begin{table}[t]
% p{3cm}<{\centering}
\centering
\footnotesize
\caption{Performance (\%) comparisons of State-of-the-art occluded ReID method and CACE-Net on Occluded Duke.} 
\label{table_perf_occluded}
\begin{tabular}{lll}
\hline
Method &mAP&Rank-1\\
\hline
PCB (ECCV2018)\cite{sun2018beyond} & 33.7 & 42.6 \\ %& 98.31\\
DSR (CVPR2018)\cite{he2018deep} & 30.4 & 40.8 \\ %& 98.31\\
SFR \cite{he2018recognizing} & 32.0 & 42.3 \\ %& 98.31\\DSR (ICCV 2019)\cite{miao2019pose} & 30.4 & 40. \\ %& 98.31\\
PGFA (ICCV 2019)\cite{miao2019pose} & 37.3 & 51.4 \\ %& 98.31\\
HOReID (CVPR 2020) \cite{wang2020high} & 43.8 & 55.1 \\ %&& \textbf{98.28}\\
SGSFA (ACML 2020) \cite{ren2020semantic} & \underline{47.4} & \textbf{62.3} \\ %&& \textbf{98.28}\\
CACE-Net &  \textbf{50.8} & \underline{58.8} \\ %& 98.22 \\
\hline

\end{tabular}
\end{table}

We evaluate our proposed CACE-Net with the state-of-the-art ReID models. These methods include: (1) the part-based models such as Pyramid, RelationNet; (2) the alignment-based methods like AlignReID, VPM, HOReID;
%(3) the human semantic parsing-based methods like SPReID,DSA-reID; 
(3) the attention-based methods like ABD-Net, RGA-SC, SCSN; (4) Joint learning methods including DCCs and SMI that learns conditional features with RNN; (5) ReID methods that utilizes graph structure such as Group-shuffling Random Walk and SGGNN. Table \ref{table_sota} shows the performance comparison of CACE-Net with State-of-the-Art methods. As shown in Table \ref{table_sota}, thanks to our novel ReID framework that integrates visual clue alignment and conditional feature embedding, CACE-Net outperforms most of the state-of-the-art methods on Market1501, DukeMTMC and MSMT-17. 

Furthermore, our CACE-Net can also be applied in ReID in occluded person. As shown in Table \ref{table_perf_occluded}, other than archieves state-of-the-art method on the general ReID datasets, CACE-Net also outperforms most of the existing occluded ReID methods. It further demonstrates CACE-Net has the ability to solve hard cases like  body occlusion.

\section{Conclusions}
This paper proposes a novel Person ReID framework that integrates both visual clue alignment and conditional feature embedding. Our proposed CACE-Net is able to automatically select crucial region pairs by correspondence attention module, and extract conditional feature embedding from the key-point pairs with a novel discrepancy-based GCN. %Instead of using a pre-defined Adjacency Matrix like traditional GCN, the relation graph is automatically predicted and dynamically adjusted during training with a novel correspondence attention module. Compared to standard graph convolution that smooths the feature of adjacent nodes, we propose a novel graph convolution that discovers the discrepancy of adjacent graph nodes. 
The experiments show the effectiveness of our model.

\bibliographystyle{ieee_fullname}
\bibliography{ref}

\end{document}